\documentclass[10pt,twocolumn,letterpaper]{article}

\usepackage{cvpr}
\usepackage{times}
\usepackage{epsfig}
\usepackage{graphicx}
\usepackage{amsmath}
\usepackage{amssymb}
\usepackage{multirow}
\usepackage{makecell}
\usepackage{authblk}
\usepackage{subcaption}

\usepackage[pagebackref=true,breaklinks=true,letterpaper=true,colorlinks,bookmarks=false]{hyperref}

\cvprfinalcopy %

\def\cvprPaperID{5536} %

\DeclareMathOperator*{\argmin}{arg\,min}
 
\ifcvprfinal\fi
\begin{document}

\title{One-Shot Domain Adaptation For Face Generation}

\author{Chao Yang\qquad Ser-Nam Lim\\Facebook AI}

\maketitle

\begin{abstract}
In this paper, we propose a framework capable of generating face images that fall into the same distribution as that of a given one-shot example. We leverage a pre-trained StyleGAN model that already learned the generic face distribution. Given the one-shot target, we develop an iterative optimization scheme that rapidly adapts the weights of the model to shift the output's high-level distribution to the target's. To generate images of the same distribution, we introduce a style-mixing technique that transfers the low-level statistics from the target to faces randomly generated with the model. With that, we are able to generate an unlimited number of faces that inherit from the distribution of both generic human faces and the one-shot example. The newly generated faces can serve as augmented training data for other downstream tasks. Such setting is appealing as it requires labeling very few, or even one example, in the target domain, which is often the case of real-world face manipulations that result from a variety of unknown and unique distributions, each with extremely low prevalence. We show the effectiveness of our one-shot approach for detecting face manipulations and compare it with other few-shot domain adaptation methods qualitatively and quantitatively.

\end{abstract}

\section{Introduction}
Deep learning has been prevailing in a variety of computer vision tasks, especially in supervised settings such as learning for classification, detection or segmentation~\cite{he2016deep,ren2015faster,liu2016ssd,he2017mask}. Deep generative models such as Variational AutoEncoder (VAE)~\cite{kingma2013auto,lombardi2018deep} and Generative Adversarial Networks (GAN)~\cite{goodfellow2014generative,olszewski2017realistic,thies2016face2face,faceswap,df_github,zollhofer2018state} in particular have gained significant prominence in the field of deep learning due to their ability to generate highly realistic images depicting faces, natural scenes and objects.

Recent advances in deep learning have paved the way for many important applications ranging from super resolution, movie making, game development, cross domain style transfer, face synthesis and aging prediction, image inpainting, photo editing and others. However, the advent of DL has also precipitated the emerging of applications that abuse its power. Technologies such as Face2Face~\cite{thies2016face2face}, FaceSwap~\cite{faceswap}, and encoder-decoder DeepFake~\cite{df_github} have resulted in the rise of online impersonations/fabrication of news, threatening even to sway the outcomes of elections.

\begin{figure}[!t]
\centering
\small
\setlength{\tabcolsep}{1pt}
\begin{tabular}{cc}
  \includegraphics[width=.24\textwidth]{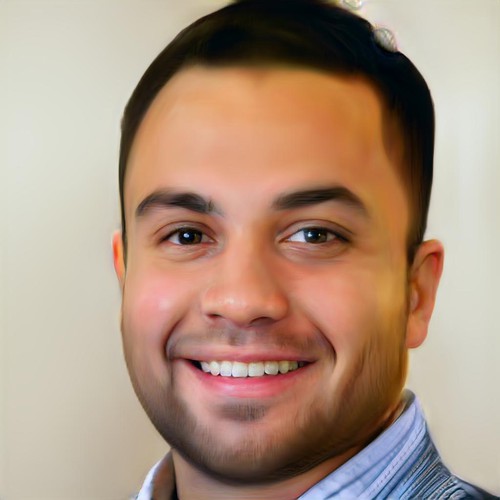} &
  \includegraphics[width=.24\textwidth]{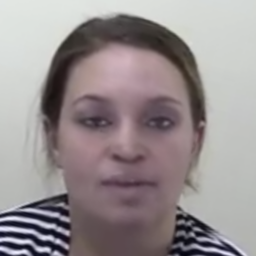} \\
  (a) & (b) \\
  \includegraphics[width=.24\textwidth]{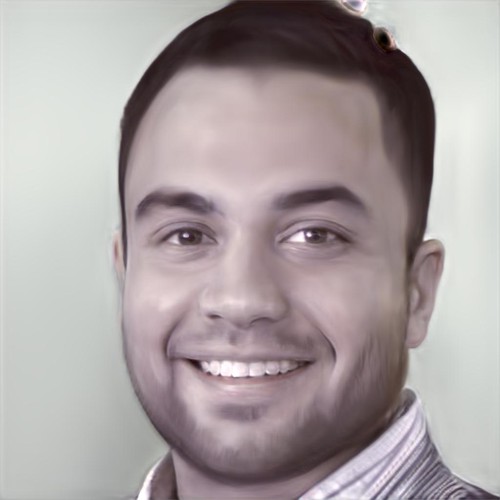} &
  \includegraphics[width=.24\textwidth]{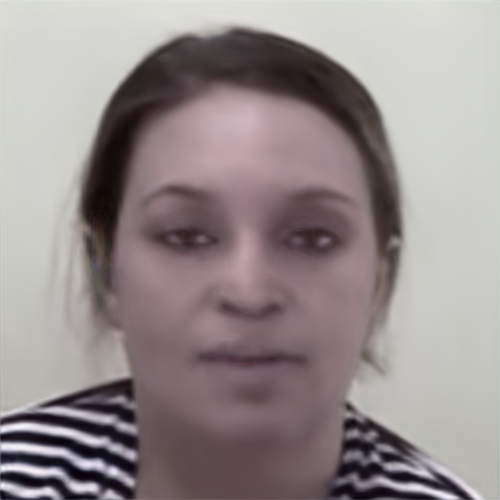} \\
  (c) & (d) \\
\end{tabular}
\caption{One-shot domain adaptation on encoder-decoder DeepFake using StyleGAN generator. (a). A Random StyleGAN generated image. (b). A one-shot image from encoder-decoder DeepFake of DFDC~\cite{dolhansky2019deepfake}. (c). A StyleGAN generated image using the same random latent input as (a) after domain adaptation. (d). The StyleGAN reconsutrcted one-shot DeepFake after domain adaptation.}
  \vspace{-5pt}
\label{fig:teaser}
\end{figure}

In this paper, we are interested in detecting Deepfakes, which refers to manipulations to replace the human face of authentic media with the face of a different person~\cite{rossler2019faceforensics++,chesney2018deep,dolhansky2019deepfake}. This is often coupled with malicious intent of defaming other people or spreading fabricated news. Image generations have been trending as a popular research topic due to the rising prevalence and public interest in Deepfake. As a result, many techniques for generating images have been proposed, each with specific probabilistic distributions or fingerprints. Often times, a new type of synthetic images emerges online but the underlying techniques are unknown, and only a few bespoke examples exist. Training a classifier to detect them poses significant challenges as we are unable to infer the complete probabilistic distribution.

To this end, our method aims to mimic a complete distribution of the target domain given just one example. While most existing domain adaptation approaches try to find a feature space in which there is maximum confusion between source and target distributions~\cite{liu2018unified,motiian2017unified}, we directly manipulate the distribution in the image space. We first train a deep generative model to learn the generic probabilistic distribution of human faces on a large collection of face images. Specifically, we adopt Style-based Generator (StyleGAN)~\cite{karras2019style} given its high capacity and superior generation quality. With the trained StyleGAN model and a single example from a specific distribution, we run iterative optimization of the input style vector to project the image to the StyleGAN distribution, followed by optimizing the model weights to minimize the projection distance and shift the StyleGAN distribution to the target's. We are then able to generate an unlimited number of random faces that are from a similar distribution as the target and yet preserve the manifold of the generic human face distributions. We further transfer low-level style from the one-shot target to the StyleGAN generated images. To do so, we notice the style vector of a given image has hierarchical structures and represents different attributes at different style layers. If we replace the style vector of randomly synthesized images with the style vector of the target at late layers, we are able to transfer the low-level statistics from the target domain to the random images we generated. We refer to this procedure as style-mixing. Combining iterative weight optimization with style-mixing, the generated images not only capture the overall probabilistic distribution of the target domain but also resemble the one-shot image in low-level appearances and details. Finally, one can then use the generated images to train a model for detecting images from the target domain.  Extensive experiment shows that our detector achieves significantly improved accuracy compared with the baseline and other state-of-the-art few-shot domain adaptation and image translation techniques. 

Our contributions can be summarized as follows:
\begin{enumerate}
	\item We introduce a novel one-shot domain adaptation framework that is effective at training a face manipulation detector given a single example of a specific face manipulation distribution. 
	\item We demonstrate that the images generated with our approach, when utilized to train a classifier, achieves superior performance in telling apart real and manipulated face images.
\end{enumerate}

\section{Related Work}

\subsection{Face Manipulation and Detection}

Only recently, a new generation of AI-based image and video synthesis algorithms have become prominent due to the development of new deep generative models such as VAEs~\cite{kingma2013auto,lombardi2018deep} and GANs~\cite{goodfellow2014generative,olszewski2017realistic}. In this paper, we mostly consider face identity manipulation methods including Face2Face~\cite{thies2016face2face}, FaceSwap~\cite{faceswap}, and encoder-decoder Deepfake~\cite{df_github}. %
Other notable face manipulation methods include audio to lip-sync~\cite{suwajanakorn2017synthesizing}, training a parameter-to-video rendering network~\cite{elor2017bringingPortraits}, synthesizing dynamic-textures with deep neural networks~\cite{nagano2018pagan}, using paired video to learn a conditional GAN~\cite{olszewski2017realistic}, or training an identity-specific celebrity network~\cite{korshunova2017fast}. There are publicly available DeepFake datasets such as FaceForensics++~\cite{rossler2019faceforensics++} and DFDC~\cite{dolhansky2019deepfake}. A comprehensive state-of-the-art report has been published by Zollhofer \emph{et al.}~\cite{zollhofer2018state}.

Our interest in this paper lies in detecting such face manipulations. Existing approaches exploit specific artifacts arising from the synthesis process, such as eye blinking~\cite{li2018ictu}, or color, texture and shape cues~\cite{de2013exposing,carvalho2015illuminant}. Li \emph{et al.}~\cite{li2018ictu} observed that DeepFake faces lack realistic eye blinking, which is utilized in a CNN/RNN model to expose DeepFake videos. Yang \emph{et al.}~\cite{yang2019exposing} utilized the inconsistency in head pose to detect fake videos. As a more generic approach, Zhou \emph{et al.}~\cite{zhou2017two} proposed a two-stream CNN for DeepFake detection. 

\subsection{Few-shot Domain Adaptation}

Overcoming the need for large training sets and improving the capability of the model to generalize from few examples have been extensively studied in recent literatures~\cite{finn2017model,liu2019few,motiian2017few,valverde2019one}. Earlier work leverages generative models of appearance that share priors across classes in a hierarchical manner~\cite{fei2006one,salakhutdinov2012one}. More recently, a new category of works emerges which focuses on using meta-learning to quickly adapt models to novel tasks~\cite{finn2017model,nichol2018first,ravi2016optimization,munkhdalai2017meta}. These methods adopt better optimization strategies during training and enhance the generalizability of the model. On the other hand,~\cite{vinyals2016matching,snell2017prototypical,sung2018learning} focuses on learning image embeddings that are better suited for few-shot learning. Similarly,~\cite{dixit2017aga,hariharan2017low,wang2018low} also propose augmenting the training set for the few-shot classification task. 

\subsection{Deep Generative Model for Image Synthesis and Disentanglement} 
Deep generative models such as GAN~\cite{goodfellow2014generative} and VAE~\cite{kingma2013auto} have been very successful in modeling natural image distributions and synthesizing realistic-looking figures. Recent advances such as WGAN~\cite{arjovsky2017wasserstein}, BigGAN~\cite{brock2018large}, Progressive GAN~\cite{karras2017progressive} and StyleGAN~\cite{karras2019style} have developed better architectures, losses and training schemes. In particular, StyleGAN~\cite{karras2019style} proposes a GAN architecture to implicitly learn hierarchical latent styles that contribute to the synthesized images. Our approach leverages StyleGAN as backbone and directly takes advantage of its expressiveness and disentanglement ability. On the other hand, several recent works aim to reverse the generation process and project an image onto latent manifold of GANs, as well as manipulating the latent code to control the output~\cite{zhu2016generative,brock2016neural,abdal2019image2stylegan,bau2018gan}. Our work is motivated to not only manipulate the latent manifold, but also adjust the model-parameter manifold to shift the whole output space given an input image.
\section{Our Approach}

We first motivate our approach. We are concerned about the scenario where we spot a single face image that is suspected to be generated (aka fake), yet we have no knowledge about the technique that produced it. Our goal is to: (1) Predict the probabilistic distribution of the target given the one-shot example; (2) Sample from the distribution to synthesize random images that are similar to the target domain and; (3) Train a classifier to detect future face images generated by the same technique. At first sight, predicting the distribution of the unknown face manipulation given one example seems ill-posed and unfeasible. We address this by learning a generic face manifold as prior, and then shift the distribution towards the target domain. 

\subsection{Overview}
Our pipeline consists of the following components:

\begin{enumerate}
\item \textbf{Face Manifold Approximation}. We learn the generic probabilistic distribution of human faces by training StyleGAN on a large collection of natural face images. All possible style vectors of the trained StyleGAN shall span a low-dimensional space that approximates the generic face manifold. 
\item \textbf{One-shot Manifold Projection}. Given a manipulated face as input, we fix the weight of the StyleGAN model and optimize the style vector to minimize the distance between the synthesized image and the input. Doing so enables us to find the one-shot's nearest neighbor on the StyleGAN manifold. In other words, we \emph{project} the target image onto the manifold.
\item \textbf{StyleGAN Manifold Shifting}. After finding the nearest neighbor of the input image, we then fix the corresponding style vector and update the StyleGAN model weights to again minimize the distance between the synthesized image and the target. Updating the weights of the model shifts the output manifold towards the target distribution. 
\item \textbf{Style Mixing}. We generate a large number of random faces from the updated StyleGAN model. Each time we generate a face, we replace the final layers of the random style vector with those of the target, such that we transfer the low-level statistics from the target to the generated images.
\item \textbf{Deepfake Detection}. We use the generated images as training data to learn model for detecting images in the target domain.
\end{enumerate}

\subsection{Face Manifold Approximation}
\label{subsec:fma}
Deep generative models are rich, hierarchical models that can learn probability distributions of the training data. As the first step, we resort to these models to learn the generic distribution of faces. We begin by training a deep generative model on a large collections of face images. If the model has sufficient capacity and is well trained, the entirety of its generated images shall span a low-dimension space that approximates the real-world face manifold. Furthermore, given enough training data, the larger capacity the model has, the more closely the output manifold would match with the true face distribution. We consider a few GAN variants including StyleGAN~\cite{karras2019style}, ProGAN~\cite{karras2017progressive} and WGAN-GP~\cite{gulrajani2017improved} as candidate models to learn the face manifold.

We analytically examine the capacity and expressiveness of the models by running the following experiment: we first train all three models on real-world face images. After the models are trained, we select one of the models as model A, and fine-tune it with images generated from model B. We then train a classifier on real vs images generated by fine-tuned A, and then test on real vs images generated by B. We can expect that, if A has higher capacity than B, it would learn to generate images with similar distributions and coverages as B. Otherwise, if B is more expressive, it is difficult for fine-tuned A to recover model B's manifold, hence the classification accuracy would be low. Table.~\ref{table:generalization} lists the classifier generalization results, which clearly shows that StyleGAN is most expressive amongst the candidate models. In addition, StyleGAN generates the most realistic and high-resolution images compared with other generative models. For these reasons, we utilize a StyleGAN model trained on an online collection of high-resolution face images as the base model for our approach.

\begin{table}[h!]
\begin{center}
  \begin{tabular}{ c | c | c}
    \hline
     \textbf{Model A} &  \textbf{Model B} & \textbf{Classification Accuracy} \\ \hline
     \multirow{2}{*}{StyleGAN} & \textbf{ProGAN} & 99.6\% \\ \cline{2-3}
                                & \textbf{WGAN-GP} & 99.4\% \\ \cline{1-3}
    \multirow{2}{*}{ProGAN} & \textbf{StyleGAN} & 72.7\% \\  \cline{2-3}
                                & \textbf{WGAN-GP} & 98.1\% \\ \cline{1-3}
    \multirow{2}{*}{WGAN-GP} & \textbf{StyleGAN} & 68.5\% \\  \cline{2-3}
                                & \textbf{ProGAN} & 88.2\% \\ \hline
  \end{tabular}
  \end{center}
  \caption{Comparing the capacity of StyleGAN, ProGAN and WGAN-GP. Higher classification accuracy indicates A has larger capacity and could better mimic the distribution of B.}
  \label{table:generalization}
\end{table}

\subsection{StyleGAN Manifold Projection}
\label{subsec:dmp}
The original StyleGAN consists of a mapping network $f$ and a synthesis network $g$. $f$ takes random noise as input and outputs a style vector $s$. $s$ is modeled as an 18 layer vector. The synthesis network takes the style vector $s$ and a random noise vector as input, and $s$ is used as parameters for adaptive instance normalization ~\cite{huang2017arbitrary} to transform the output after each convolution layer. Karras \emph{et al.}~\cite{karras2017progressive} shows that using style vector as layer-wise guidance not only makes synthesizing high-resolution images easier, but also leads to hierarchical disentanglement of local and global attributes. For our purpose, we consider a trained StyleGAN model. In this case, all possible style vectors generated by the mapping network form a synthetic-face manifold that mimics the true distribution of human faces.

With the StyleGAN manifold and a visual example from an arbitrary distribution, our next step is to project the example onto the manifold. To do so, we first detect the facial landmarks and preprocess the image by cropping it to be $1.3$ times larger than the face region, followed by resizing it to 1024x1024 which is the output size of StyleGAN. Let the preprocessed image be $I$. Projecting $I$ to StyleGAN manifold means we would like to find the style vector $s_I$ that the generated image $g(s_I)$ is most similar to $I$. In this way, $s_I$ is the style vector corresponding to $I$'s manifold projection. This process could be more formally formulated as solving for the following objective function:
\begin{equation}
s_I = \argmin\limits_{s} D(g(s), I).
\label{eqn:obj}
\end{equation}
With a differentiable distance function $D$, we can solve for Eqn.~\ref{eqn:obj} by backpropagating the loss $D$ through $g$ with the weights fixed, and then iteratively update $s$ until the loss converges. This is similar to fine-tuning using the given example $I$, but here we are optimizing $s$ instead of the weights of $g$. It is also important to use an appropriate distance function $D$. Common candidates for reconstruction loss are $\ell_1$, $\ell_2$ and CNN-based perceptual loss. We experimented with those losses and found that using a combination of perceptual and $\ell_1$ loss leads to the best visual quality and reconstruction fidelity:
\begin{equation}
D(g(s), I) = \sum\limits_{l} \|f_l(g(s))-f_l(I)\|_2^2 + \lambda\|g(s)-I\|_1^1.
\label{eqn:dis}
\end{equation}
Here $f_l$ is the neuron responses at $l_{th}$ layer extracted with a pre-trained VGG-16 model, and $\lambda=5$ is the weight of $\ell_1$ loss. The reconstruction loss usually converges within 1,000 iterations of optimization. After it converges, the style vector $s_I$ is taken as the projection of $I$ on the StyleGAN manifold, and the reconstruction $g(s_I)$ is the nearest neighbor of $I$ amongst StyleGAN output images.

A more accurate projection requires optimizing the style vector and the noise vector at the same time. However, we found that the noise vector had little effects on the final reconstruction output. In our experiments, we always initialize the style vector to be a zero vector and the noise vector to be random Gaussian, and we update the style vector but keep the noise fixed during optimization.

\subsection{StyleGAN Manifold Shifting}

After we found the projection of the target on the original StyleGAN manifold, our next step is to shift the StyleGAN manifold towards the target distribution. To do so, we use similar iterative optimization procedure as~\ref{subsec:dmp}. However, instead of updating the style vector $s$, we fix $s$ to be the output of~\ref{subsec:dmp} $s_I$ while updating the model weights to match the generated image with the target. The idea here is that every time we update the weight of $g$, we are slightly adjusting the StyleGAN manifold when the weight changes are sufficiently small. By fixing the style vector to be $s_I$ and updating the model weights of $g$, we are pulling the nearest neighbor of StyleGAN manifold closer to the target such that the entire manifold becomes more similar to the target distribution. Similar to Eqn.~\ref{eqn:obj}, the objective function can be defined as:
\begin{equation}
g_I = \argmin\limits_{g} D(g(s_I), I).
\label{eqn:obj2}
\end{equation}
Here, we reuse Eqn.~\ref{eqn:dis} as the distance function. As far as the optimization is concerned, it comes down to the choice of updating different layers of the StyleGAN. As shown in~\cite{karras2019style}, the late layers of the style vector control the low-level details of the output image such as the color or local textures, while the initial layers control the global attributes such as gender, appearance or identity. We experimented updating different StyleGAN layers for manifold shifting, and examine the synthetic image quality and the domain adaptation effectiveness. Our observation is that updating all StyleGAN layers makes the optimized model generate images most similar to the target and also achieve the highest accuracy when being used to train classifiers. Since we already inferred $s_I$ that generates an image similar to $I$, this step would only slightly adjust the weights of the model. In this case, the optimized model still preserves the generic face manifold learned in~\ref{subsec:fma}. 

\begin{figure}[!h]
\centering
\small
\setlength{\tabcolsep}{1pt}
\begin{tabular}{ccc}
  \includegraphics[width=.15\textwidth]{figures/qual_results_new/ae/autoencoder.png} &
  \includegraphics[width=.15\textwidth]{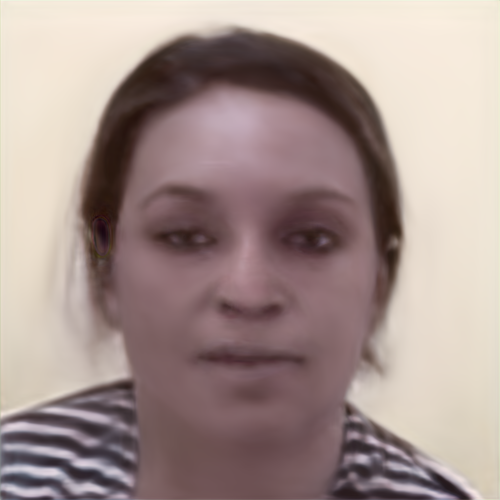} &
  \includegraphics[width=.15\textwidth]{figures/qual_results_new/ae/reconstructed.png} \\
   \includegraphics[width=.15\textwidth]{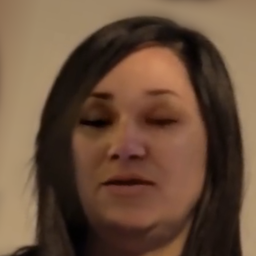} &
  \includegraphics[width=.15\textwidth]{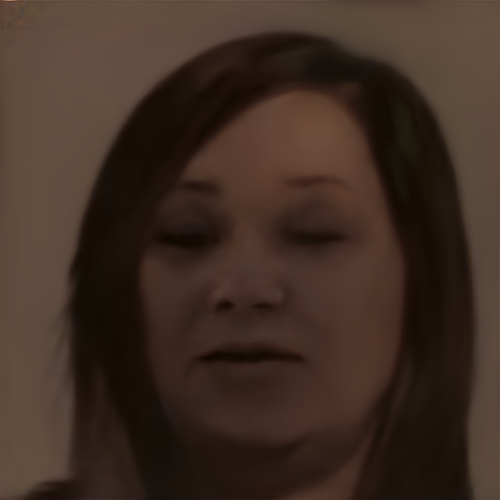} &
  \includegraphics[width=.15\textwidth]{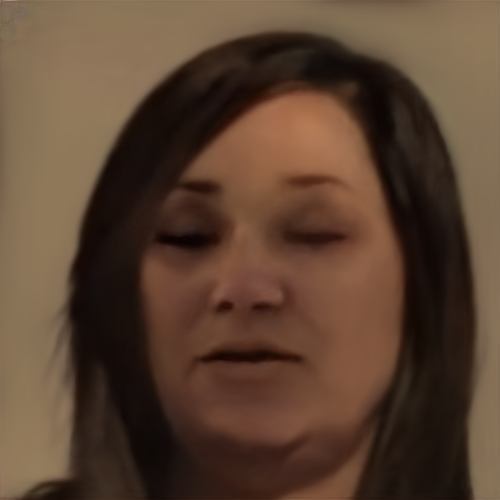} \\
   \includegraphics[width=.15\textwidth]{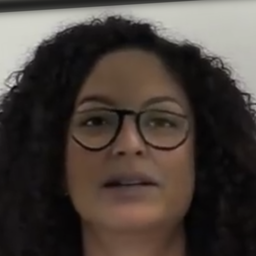} &
  \includegraphics[width=.15\textwidth]{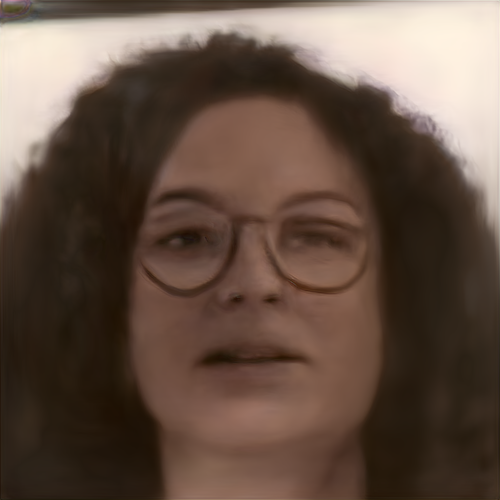} &
  \includegraphics[width=.15\textwidth]{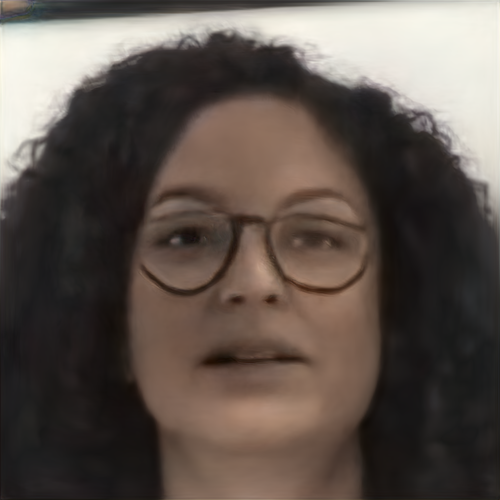} \\
\end{tabular}
\caption{The reconstruction after manifold projection and manifold shifting for one-shot encoder-decoder (top), neural talking head (middle) and FSGAN (bottom). From left to right: the input image; $g(s_I)$ as the reconstructed image after manifold projection; $g_I(s_I)$ as the reconstructed image after manifold shifting.}
\label{fig:before_after}
  \vspace{-5pt}
\end{figure}

To better illustrate the effects of manifold projection and shifting, Fig.~\ref{fig:before_after} shows visual examples of an input, the reconstructed image after manifold projection, and the reconstructed image after manifold shifting. It shows that after manifold shifting, the reconstructed image matches with the input more closely in global color and appearances.

\subsection{Style Mixing}

In the previous steps, we optimize the weights of the StyleGAN model such that it generates images of similar distribution as the target. However, it does not suffice to simply change the global appearances as there are certain low-level statistics that the target exhibits, which are useful signals when training a classifier. We propose to use style mixing to further generate images that match with the target in low-level details. Karras \emph{et al.}~\cite{karras2019style} shows that StyleGAN's style vector comes with the property of disentanglement, which separates the high-level and low-level attributes of the synthesized image at different style layers. Based on this, we use the style vector $s_I$ inferred in~\ref{subsec:dmp} as the interpretable representation of the target. For each random style vector $s$ that we sampled with the mapping network, we replace the final layers of $s$ with those of $s_I$ before giving it as input to the generator so that the generated random image $g(s)$ inherits the low-level color and textures from $I$.

We experimented with replacing different number of the final layers, and found that replacing the three last layers of $s$ with those of $s_I$ preserves the global appearance of the image yet still manages to change the output to more closely resemble $I$. Combining manifold shifting and style mixing, the generated images not only capture the generic human face manifold but also display low-level statistics of the target. 

\subsection{Classification}
The last step is to use the randomly generated images as a synthetic dataset to train a classifier against the target domain. In the case of face manipulation detection, we train a classifier between real images and one-shot optimized StyleGAN synthetic images, and use it to detect the actual face manipulation images from the real images. Another task we could solve using the synthetic datasets is multi-domain classification, where given an image we could classify it into a specific type of face manipulations. For all classification tasks, we use ResNet50~\cite{he2016deep} as the backbone model.

\section{Experiments}
\label{sec:setup}
\subsection{Experiment Setup and Results}
We evaluate our method on several face manipulation algorithms to show its effectiveness. We use DFDC~\cite{dolhansky2019deepfake} and FaceForensics++~\cite{rossler2019faceforensics++}, which consists of a large number of videos that are generated by different face manipulation techniques including encoder-decoder Deepfake~\cite{df_github}, Face2Face~\cite{thies2016face2face} and FaceSwap~\cite{faceswap}, neural talking head~\cite{zakharov2019few} and FSGAN~\cite{nirkin2019fsgan}. For each of the algorithms, we randomly select one image from the dataset. We then apply our one-shot domain adaptation to shift the StyleGAN distribution towards the image and mix the low-level styles and generate a large number of random faces. Finally, we train a classifier using the generated faces to detect images in the target domain.

\begin{figure*}[!h]
\centering
\small
\setlength{\tabcolsep}{1pt}
\begin{tabular}{ccccc}
  \includegraphics[width=.19\textwidth]{figures/qual_results_new/ae/autoencoder.png} &
  \includegraphics[width=.19\textwidth]{figures/qual_results_new/ae/reconstructed.png} &
  \includegraphics[width=.19\textwidth]{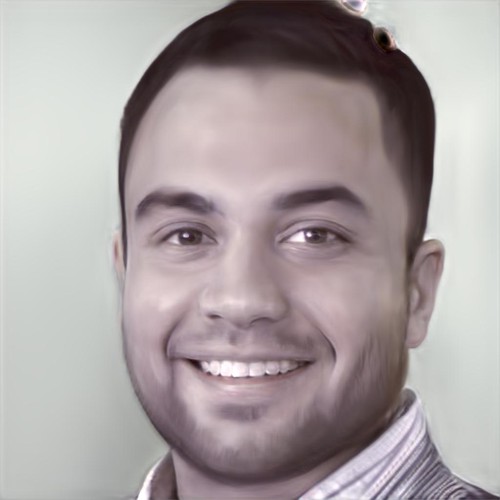} &
   \includegraphics[width=.19\textwidth]{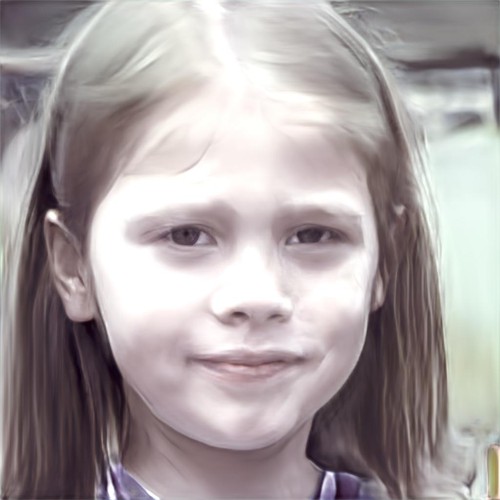}& 
   \includegraphics[width=.19\textwidth]{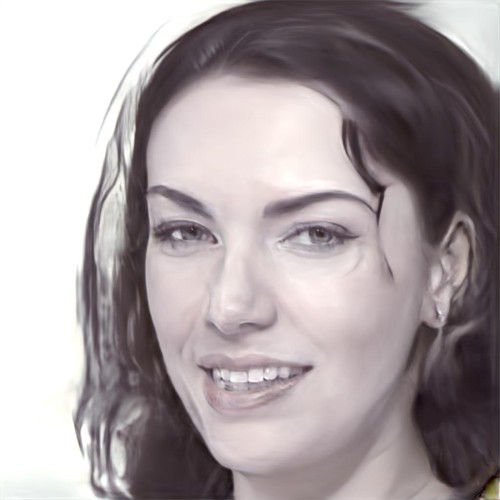} \\
    \includegraphics[width=.19\textwidth]{figures/qual_results_new/nh2/rotated.png} &
  \includegraphics[width=.19\textwidth]{figures/qual_results_new/nh2/regenerated.png} &
  \includegraphics[width=.19\textwidth]{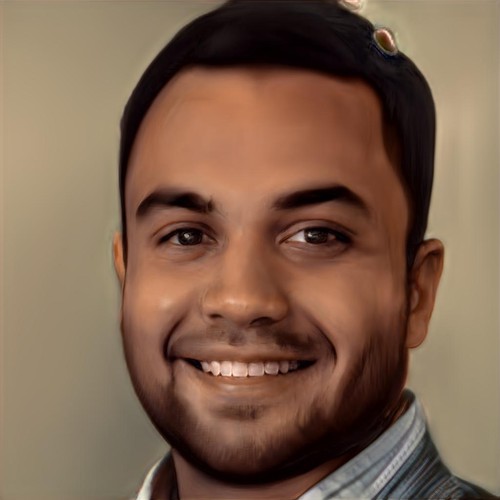} &
   \includegraphics[width=.19\textwidth]{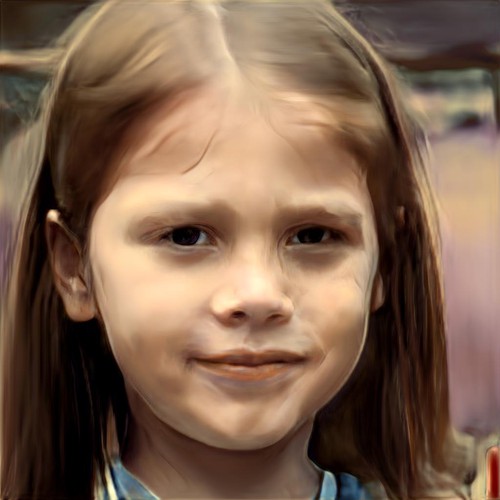}& 
   \includegraphics[width=.19\textwidth]{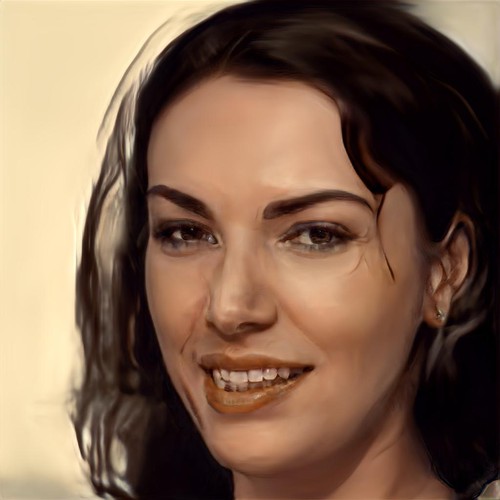} \\
    \includegraphics[width=.19\textwidth]{figures/qual_results_new/fsgan2/rotated.png} &
  \includegraphics[width=.19\textwidth]{figures/qual_results_new/fsgan2/regenerated.png} &
  \includegraphics[width=.19\textwidth]{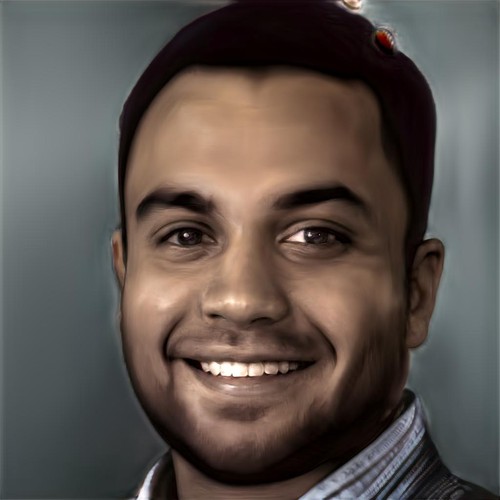} &
   \includegraphics[width=.19\textwidth]{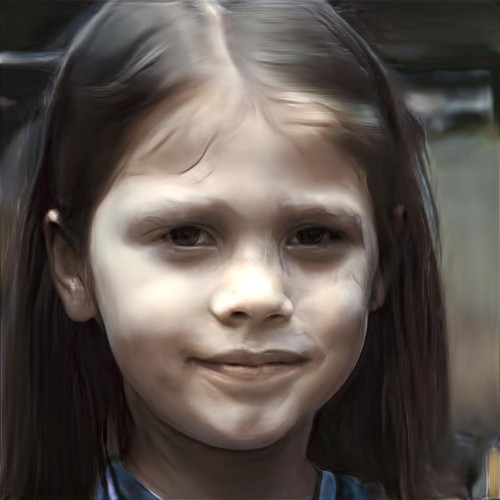}& 
   \includegraphics[width=.19\textwidth]{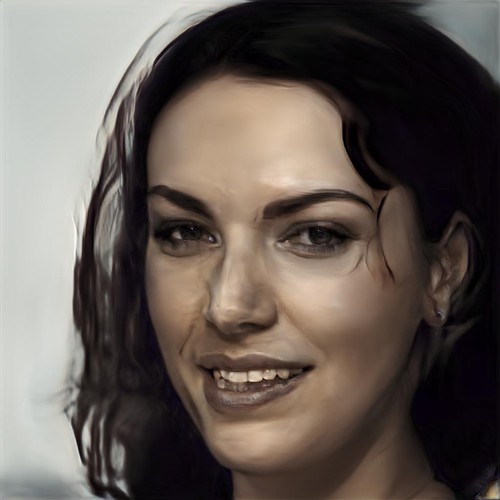} \\
   (a) Target & (b) Reconstructed image & \multicolumn{3}{c}{(c) Randomly sampled images.}
\end{tabular}
\caption{Visual examples of one-shot DeepFake domain adaptation using StyleGAN. From top to bottom: encoder-decoder, neural talking head and FSGAN. }
\vspace{-5pt}
\label{fig:visual}
\end{figure*}

Qualitatively, we show visual results of one-shot domain adaptation in Fig.~\ref{fig:visual}. For each dataset, we show the one-shot image, the reconstructed image corresponding to the target, and randomly sampled images that mimic the target distribution. At each column, we use the same random style vector to generate the images so that they have the same identities. However, their appearances change according to the one-shot input. We can see that the reconstructed image, which is the closest neighbor of the target on the shifted StyleGAN manifold, visually resembles the target. The randomly sampled images also inherit similar appearances and low-level characteristics from the target.

For quantitative evaluation, we first generate 10,000 random images using the one-shot example for each of the face manipulation techniques. We then train a classifier using real face images as real and the 10,000 synthetic images as fake, and then test on real face images vs the actual face manipulation images (encoder-decoder, Face2Face or FaceSwap). As the baseline, we train the classifier using real faces images and 10,000 random images generated by the original StyleGAN (without domain adaptation) and evaluate on real face images vs actual face manipulations. Table.~\ref{table:quant} shows the results. From it we can see that without domain adaptation, the detection accuracy is low. This is expected as the original StyleGAN generated images have a distinct distribution from the target. However, after domain adaptation using the one-shot example, the classification accuracy improves significantly - almost perfect for all three datasets. This shows that our one-shot domain adaptation is effective at generating images with similar distribution to the target domain, at the slight cost of seeing one more image compared with the baseline. In addition to train a binary classifier on real vs manipulated, we further experimented with fine-grained classification by training a multi-domain classifier using all images from the three StyleGAN-synthetic face manipulation domains. The high accuracy of fine-grained classification (82.1\%) shows that the our synthetic datasets have distinguishable distributions from each other and their distributions are also consistent with the target domain.

\begin{table*}[h!]
\begin{center}
  \begin{tabular}{ c  c  c}
    \hline
    \textbf{Train} & \textbf{Test} &  \textbf{Average Precision} \\ \hline
     Real/StyleGAN & Real/encoder-decoder & 35.2\% \\
     Real/one-shot encoder-decoder & Real/encoder-decoder & 93.4\% \\ \hline
     Real/StyleGAN & Real/Face2Face &  35.3\% \\
     Real/one-shot Face2Face & Real/Face2Face &  99.2 \% \\ \hline
     Real/StyleGAN & Real/FaceSwap & 41.6\% \\
     Real/one-shot FaceSwap & Real/FaceSwap & 95.2\% \\ \hline
     Real/(one-shot) encoder-decoder/Face2Face/FaceSwap & Real/encoder-decoder/Face2Face/FaceSwap & 82.1\% \\
    \hline
  \end{tabular}
  \end{center}
  \caption{Quantitative evaluation results. (One-shot) encoder-decoder Deepfake/Face2Face/FaceSwap are the synthetic datasets generated by StyleGAN after running domain adaptation algorithm given the one-shot example of encoder-decoder Deepfake/Face2Face/FaceSwap (Fig.~\ref{fig:visual} (a)). }
  \vspace{-5pt}
  \label{table:quant}
\end{table*}

\begin{figure}[!h]
\centering
\small
\setlength{\tabcolsep}{1pt}
\begin{tabular}{cc}
 \includegraphics[width=.24\textwidth]{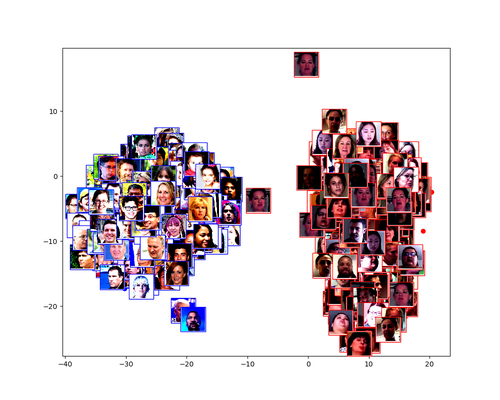} &
   \includegraphics[width=.24\textwidth]{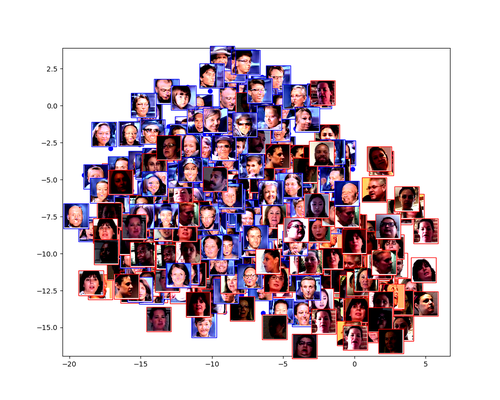} \\
\end{tabular}
\caption{t-SNE embedding visualizations. Left: Embeddings of original StyleGAN generated images and encoder-decoder Deepfake images. Right: Embeddings of one-shot domain-adapted StyleGAN generated images and Deepfake images.}
  \vspace{-5pt}
\label{fig:t_sne}
\end{figure}

In Fig.~\ref{fig:t_sne}, we plot the t-SNE embeddings of StyleGAN generated image before and after domain adaptation, comparing with the embeddings of actual fake images. We can see that before domain adaptation, the embeddings of StyleGAN generated images and encoder-decoder Deepfake images are separated. After domain adaptation, the embeddings between the two domains are much closer to each other.

\subsection{Ablation Study}

\begin{figure}[!h]
\centering
\small
\setlength{\tabcolsep}{1pt}
\begin{tabular}{cccc}
\includegraphics[width=.11\textwidth]{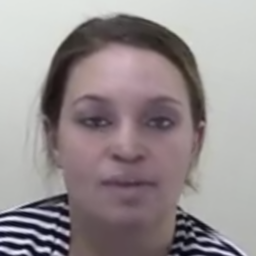} &
 \includegraphics[width=.11\textwidth]{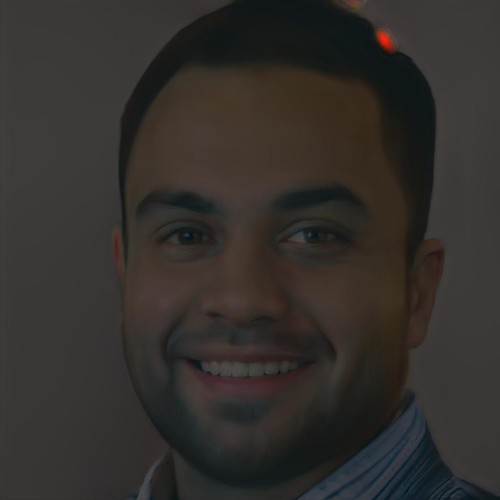} &
\includegraphics[width=.11\textwidth]{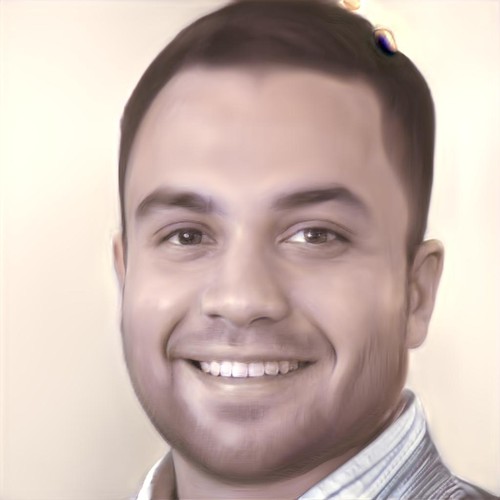} &
\includegraphics[width=.11\textwidth]{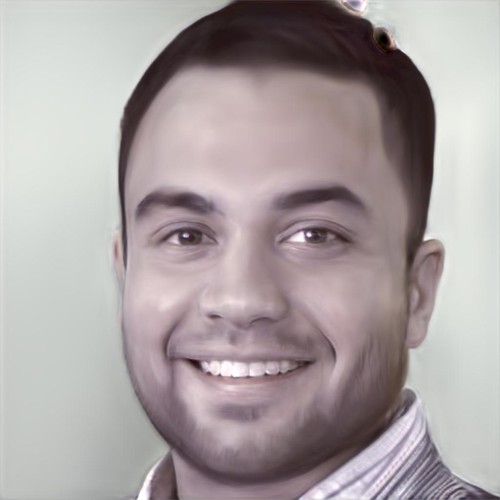} \\
\end{tabular}
\caption{Analyzing the effects of different components. From left to right: Target encoder-decoder Deepfake image; Randomly generated image after manifold shifting without style mixing; Randomly generated image using original StyleGAN with style mixing; Randomly generated image after manifold shifting and style mixing.}
  \vspace{-5pt}
\label{fig:comp_analyze}
\end{figure}

As described, there are two main components of our approach: StyleGAN manifold shifting and style mixing. To better understand the effects of each component, in Fig.~\ref{fig:comp_analyze} we compare visual examples of a randomly generated image with manifold shifting only and with style mixing only, given the one-shot example. We can see that with only manifold shifting, the output images are different from the input in low-level characteristics. Meanwhile, if we only mix the styles without adjusting the original StyleGAN model, the output image's colors and textures are distorted and do not look realistic. In both cases, the classification accuracy using the randomly generated images significantly drops compared with the results of utilizing both components (Table. ~\ref{table:comp_analyze}). 

\begin{table}[h!]
\begin{center}
\setlength{\tabcolsep}{1pt}
  \begin{tabular}{c  c}
    \hline
    \textbf{Setting} &  \textbf{Average Precision} \\ \hline
    StyleGAN manifold shifting only & 43.1\% \\
    StyleGAN mix style only & 34.0\% \\  \hline
    Ours & 93.4\% \\ \hline
  \end{tabular}
  \end{center}
  \caption{Real vs encoder-decoder Deepfake classification results. The classification models are trained using images generated by StyleGAN with manifold-shifting only or style-mixing only.}
    \vspace{-5pt}
  \label{table:comp_analyze}
\end{table}

\begin{figure}[!h]
\centering
\small
\setlength{\tabcolsep}{1pt}
\begin{tabular}{ccccc}
  \includegraphics[width=.09\textwidth]{figures/ablation/autoencoder.png} &
  \includegraphics[width=.09\textwidth]{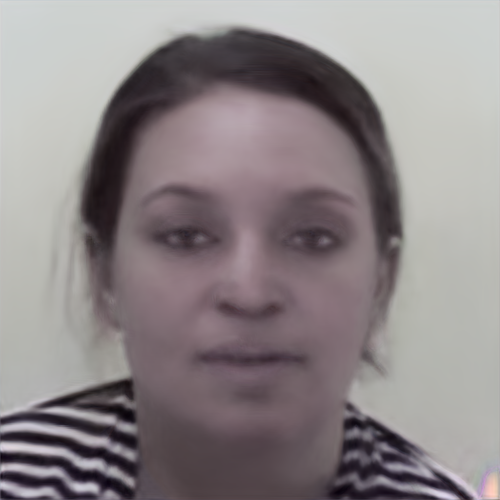} &
  \includegraphics[width=.09\textwidth]{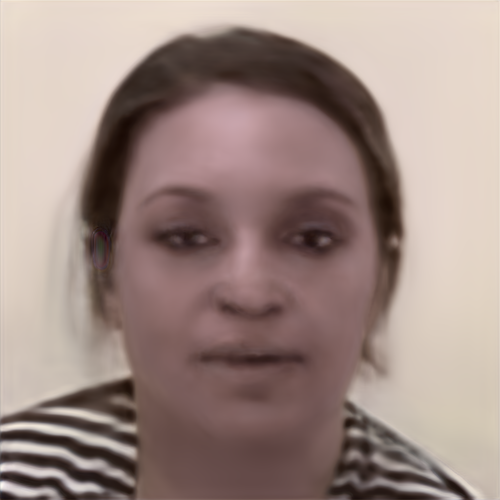} &
  \includegraphics[width=.09\textwidth]{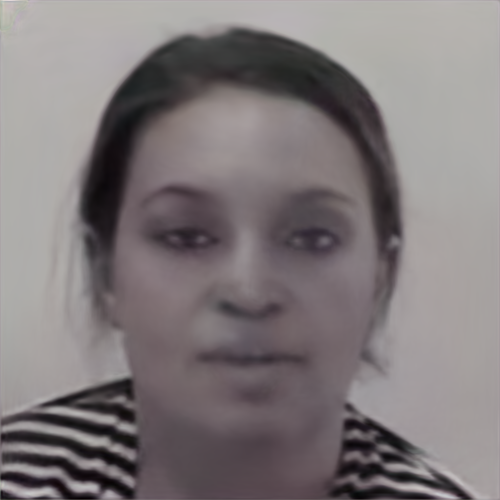} &
  \includegraphics[width=.09\textwidth]{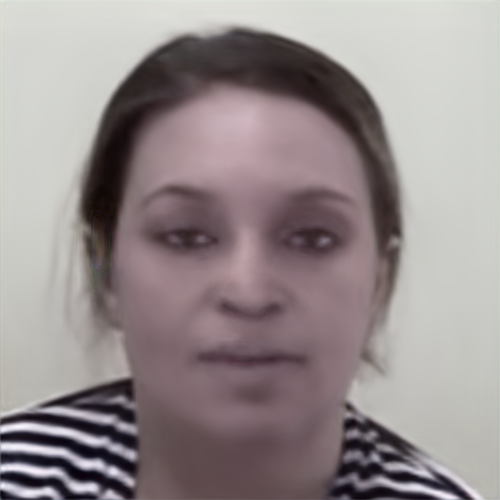} \\
\end{tabular}
\caption{Input reconstruction after optimizing style and weights using different losses. From left to right: input, $\ell_1$ loss, $\ell_2$ loss, VGG-16 loss, combining $\ell_1$ and VGG-16 losses.}
  \vspace{-5pt}
\label{fig:loss_type}
\end{figure}

As for the reconstruction losses used for optimizing the style vector and the model weights, we experimented with $\ell_1$, $\ell_2$, VGG-16, and a combination of these losses. We observe that the reconstruction quality is correlated with the reconstruction loss used. As shown by the example images in Fig.~\ref{fig:loss_type}, using a single loss of either $\ell_1$, $\ell_2$ or VGG-16 results in bleached color or distorted appearances compared with accurate reconstructions generated by using a combination of $\ell_1$ and perceptual loss. Note that for all VGG-16 losses, we use all the layers to compute the feature responses when measuring the perceptual similarity.

\begin{figure}[!h]
\centering
\small
\setlength{\tabcolsep}{1pt}
\begin{tabular}{cccc}
  \includegraphics[width=.12\textwidth]{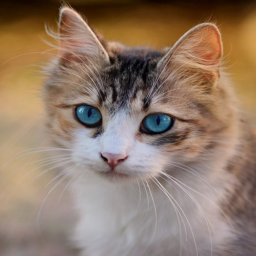} &
  \includegraphics[width=.12\textwidth]{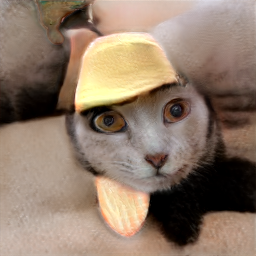} &
  \includegraphics[width=.12\textwidth]{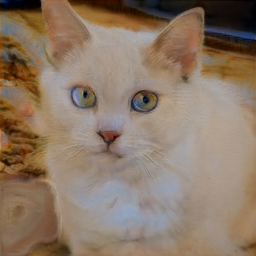} &
  \includegraphics[width=.12\textwidth]{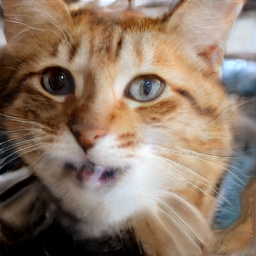} \\
  \includegraphics[width=.12\textwidth]{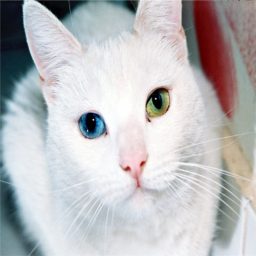} &
  \includegraphics[width=.12\textwidth]{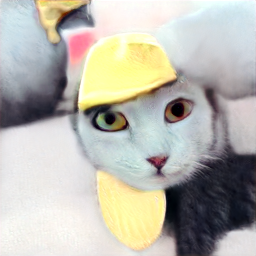} &
  \includegraphics[width=.12\textwidth]{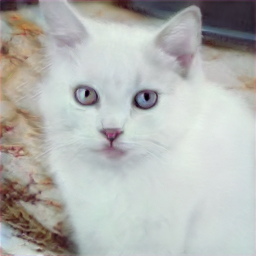} &
  \includegraphics[width=.12\textwidth]{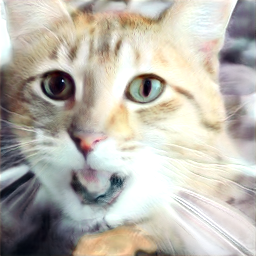} \\
  \includegraphics[width=.12\textwidth]{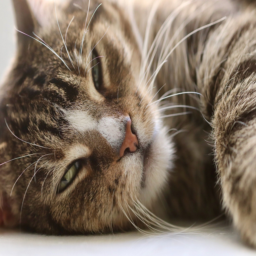} &
  \includegraphics[width=.12\textwidth]{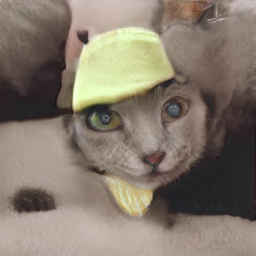} &
  \includegraphics[width=.12\textwidth]{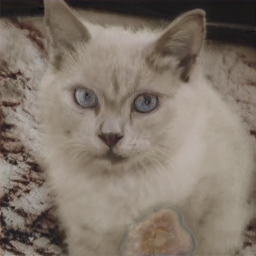} &
  \includegraphics[width=.12\textwidth]{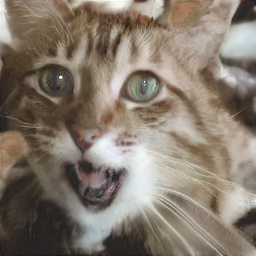} \\
  (a) Input & \multicolumn{3}{c}{(b) Randomly generated images}\\
\end{tabular}
\caption{One-shot domain adaptation on cats.}
  \vspace{-5pt}
\label{fig:cat_results}
\end{figure}

To show the effectiveness of our approach on datasets other than face manipulations and that it can be used as a generic domain adaptation approach, we further show that our one-shot domain adaptation technique can be applied to other domains such as cats. Given a single cat image as input and a pre-trained StyleGAN cat model, we can generate random images that are visually similar to the target (Fig.~\ref{fig:cat_results}).

\subsection{Comparisons}

\noindent\textbf{Comparison with Few-shot Classification}
We compare our results with few-shot classifications. We train a classifier using different number of examples from the target domain (encoder-decoder Deepfake) and use it to classify the target from the real faces. Table.~\ref{table:few_shots_clf} shows the results. We can see that directly training a classifier with very few examples (1 or 10) in the target domain leads to inferior performance compared with ours. Only when the number of examples from the target domain is large enough (over 100), the classifier can achieve high accuracy. For all the experiments, we use 10,000 real face images as real, and adjust the weight between false positive loss and false negative loss to reflect the imbalanced quantity of real and fake.

\begin{table}[h!]
\begin{center}
\setlength{\tabcolsep}{1pt}
  \begin{tabular}{c c c} 
    \hline
    \textbf{Train} & \textbf{Test} & \textbf{Average Precision} \\ \hline
    Real/1-shot DF Clf & Real/DF & 52.1\% \\ 
    Real/10-shot DF Clf & Real/DF & 79.7\% \\ 
    Real/100-shot DF Clf  & Real/DF & 93.0\% \\ 
    Real/1000-shot DF Clf & Real/DF & 99.5\% \\  \hline
     \multicolumn{2}{c}{Ours} & 93.4 \\ \hline

  \end{tabular}
  \end{center}
  \caption{Comparing with training a encoder-decoder Deepfake classifier with different number of examples. Our one-shot domain adaptation can achieve detection accuracy on par with a 100-shot DF classifier.}
  \label{table:few_shots_clf}
    \vspace{-5pt}
\end{table}

\noindent\textbf{Comparison with Fine-tuning}
Another possibility is to fine-tune a pre-trained StyleGAN model, by further training it with a few examples from the target domain. Ideally the fine-tuned model would generate synthetic images with a similar distribution as the target domain. However, we found that fine-tuning StyleGAN with only a few examples is difficult, as the model would collapse and keep generating identical images. The classification accuracy of Table~\ref{table:few_shots} shows that only when we have sufficient examples (over 100) to fine-tune the original StyleGAN model, would it not lead to mode collapse and achieve reasonable classification accuracy. Note here we use the ProGAN (pre-trained on real face images) synthetic images as the target domain, which is more difficult to differentiate with real.

\begin{table}[h!]
\begin{center}
\setlength{\tabcolsep}{1pt}
  \begin{tabular}{c c c} 
    \hline
    \textbf{Train} & \textbf{Test} & \textbf{Average Precision} \\ \hline
    Real/1-shot ProGAN & Real/ProGAN & 10.2\% \\ 
    Real/10-shot ProGAN & Real/ProGAN & 21.8\% \\ 
    Real/100-shot ProGAN & Real/ProGAN & 88.7\% \\ 
    Real/1000-shot ProGAN & Real/ProGAN & 99.0\% \\ \hline
    \multicolumn{2}{c}{Ours} & 62.1\% \\
    \hline
  \end{tabular}
  \end{center}
  \caption{Comparing with fine-tuning StyleGAN with different number of examples.}
  \label{table:few_shots}
    \vspace{-5pt}
\end{table}

\begin{figure}[tbp]
\begin{minipage}[b]{1.0\linewidth}
\centering
\small
\setlength{\tabcolsep}{1pt}
\begin{tabular}{ccccc}
 &
  \includegraphics[width=.18\textwidth]{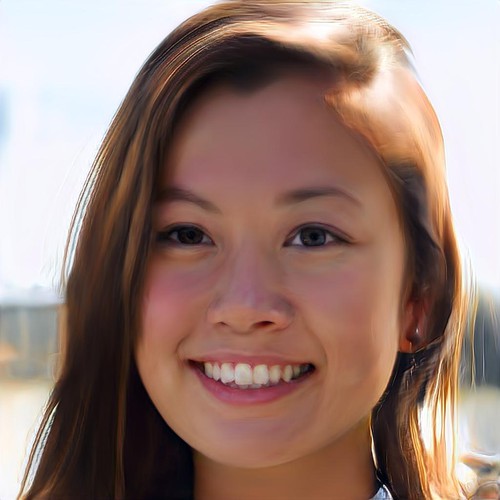} &
   \includegraphics[width=.18\textwidth]{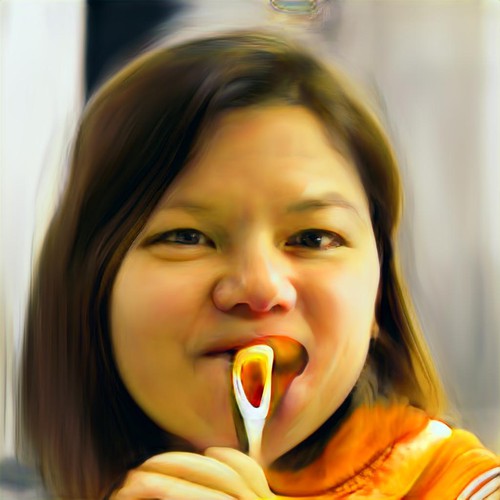} &
   \includegraphics[width=.18\textwidth]{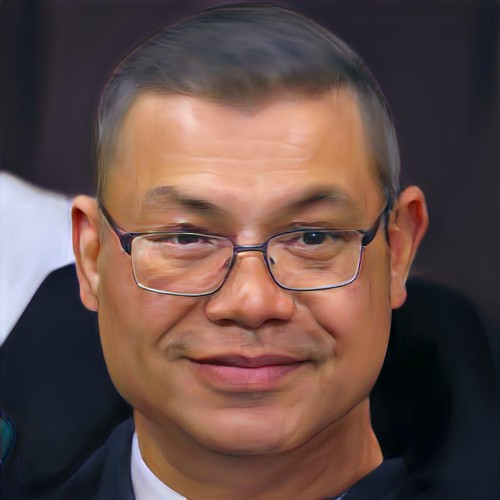} &

   \includegraphics[width=.18\textwidth]{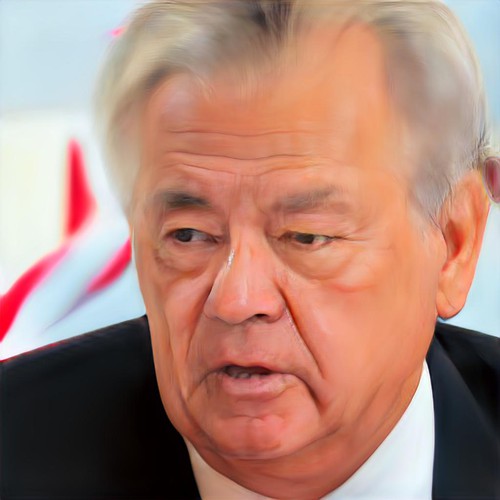}  \\
    \includegraphics[width=.18\textwidth]{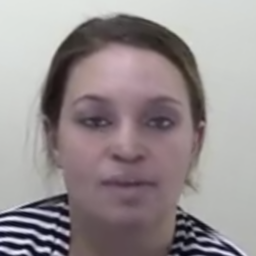} 
  &
  \includegraphics[width=.18\textwidth]{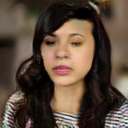} &
  \includegraphics[width=.18\textwidth]{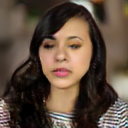} &
  \includegraphics[width=.18\textwidth]{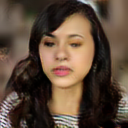} &
  \includegraphics[width=.18\textwidth]{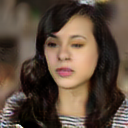} \\
  & 
  \includegraphics[width=.18\textwidth]{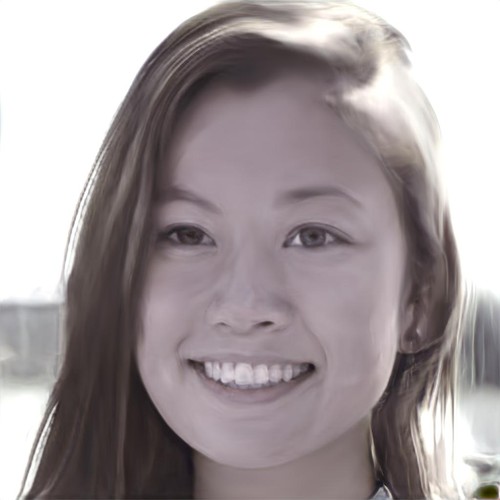} &
   \includegraphics[width=.18\textwidth]{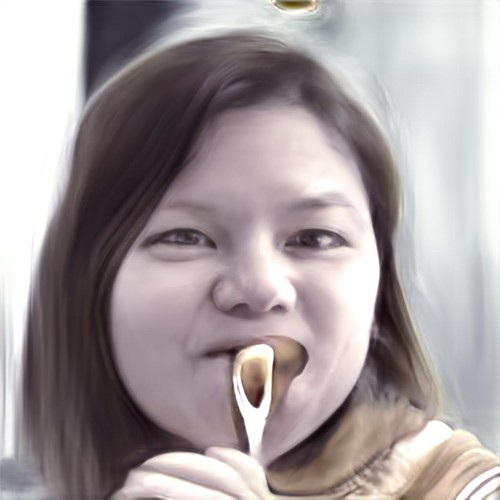} &
   \includegraphics[width=.18\textwidth]{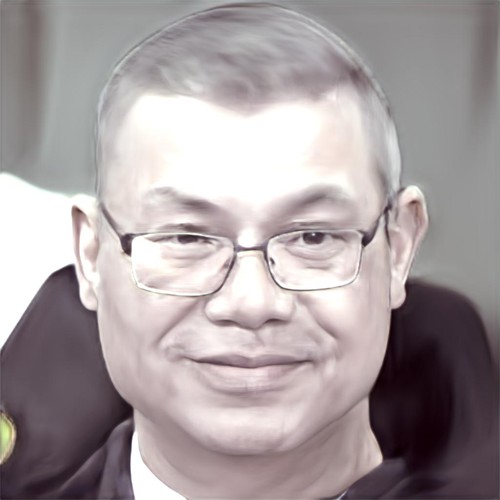} &

   \includegraphics[width=.18\textwidth]{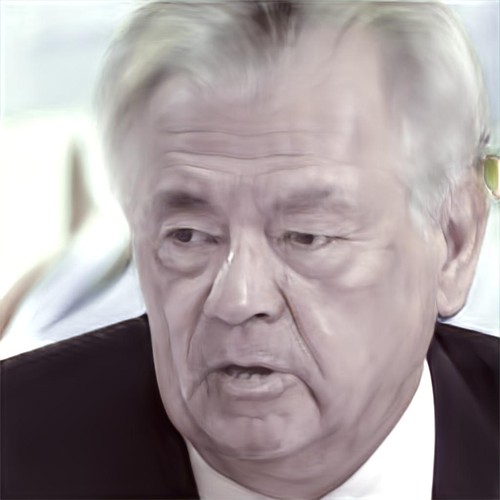}  \\
\end{tabular}
    \\
    \setlength{\tabcolsep}{5pt}
    \begin{tabular}[b]{c | c  c}
     \textbf{Method} & FUNIT & Ours \\ \hline
     \textbf{Accuracy} & 34.4\% & 93.4\% \\ %
    \end{tabular}
\end{minipage}
\caption{Above from top to bottom: Random StyleGAN generated images; Input 1-shot encoder-decoder Deepfake and translated images using FUNIT; Translated images using our one-shot domain adaptation. Below: DeepFake classification accuracy when trained on FUNIT vs our generated images.}
\label{fig:funit_results}
\end{figure}

\noindent\textbf{Comparison with FUNIT}
We also compare our results with Few-Shot Unsupervised Image-to-Image Translation (FUNIT)~\cite{liu2019few}. FUNIT could also translate an image to the target domain given a few examples of the target domain. However, at training time, FUNIT would require a large number of labeled images in over 100 classes. In contrast, our approach is entirely unsupervised and only requires pre-training the StyleGAN model. At test time, FUNIT can also translate the images from the source domain (e.g. StyleGAN synthetic images) to the target domain (e.g. encoder-decoder Deepfake) given one example. However in terms of translation quality, from Fig.~\ref{fig:funit_results} we can see that FUNIT actually modifies the identity of the source image instead of changing the appearances or low-level statistics. We further use these images to train a real/Deepfake classifier: we first use random StyleGAN generated images as the content and the 1-shot Deepfake image as the style, which are given to the trained FUNIT model to generate a synthetic dataset that adapts StyleGAN to Deepfake. We then train a classifier using real vs FUNIT translated images, and test on real vs actual Deepfake images. This results in an average precision significantly lower than ours.

\section{Conclusions}
We propose a simple yet effective one-shot domain adaptation method based on StyleGAN. Our approach not only generates compelling visual results similar to the one-shot target, but also allows us to train robust classifiers in response to different target domains. This process is also fully automatic, requiring little supervision. As future work, we would like to extend our framework to be a more generic image translation and domain adaptation approach.

{\small
\bibliographystyle{ieee}
\bibliography{egbib}

\begin{thebibliography}{10}\itemsep=-1pt

\bibitem{df_github}
Deepfakes github.
\newblock \url{https://github.com/ deepfakes/faceswap.}
\newblock Accessed: 2019-11-05.

\bibitem{faceswap}
Faceswap.
\newblock \url{https://github.com/MarekKowalski/FaceSwap/}.
\newblock Accessed: 2019-11-05.

\bibitem{abdal2019image2stylegan}
R.~Abdal, Y.~Qin, and P.~Wonka.
\newblock Image2stylegan: How to embed images into the stylegan latent space?
\newblock In {\em Proceedings of the IEEE International Conference on Computer
  Vision}, pages 4432--4441, 2019.

\bibitem{arjovsky2017wasserstein}
M.~Arjovsky, S.~Chintala, and L.~Bottou.
\newblock Wasserstein gan.
\newblock {\em arXiv preprint arXiv:1701.07875}, 2017.

\bibitem{elor2017bringingPortraits}
H.~Averbuch-Elor, D.~Cohen-Or, J.~Kopf, and M.~F. Cohen.
\newblock Bringing portraits to life.
\newblock {\em ACM Transactions on Graphics (Proceeding of SIGGRAPH Asia
  2017)}, 36(6):196, 2017.

\bibitem{bau2018gan}
D.~Bau, J.-Y. Zhu, H.~Strobelt, B.~Zhou, J.~B. Tenenbaum, W.~T. Freeman, and
  A.~Torralba.
\newblock Gan dissection: Visualizing and understanding generative adversarial
  networks.
\newblock {\em arXiv preprint arXiv:1811.10597}, 2018.

\bibitem{brock2018large}
A.~Brock, J.~Donahue, and K.~Simonyan.
\newblock Large scale gan training for high fidelity natural image synthesis.
\newblock {\em arXiv preprint arXiv:1809.11096}, 2018.

\bibitem{brock2016neural}
A.~Brock, T.~Lim, J.~M. Ritchie, and N.~Weston.
\newblock Neural photo editing with introspective adversarial networks.
\newblock {\em arXiv preprint arXiv:1609.07093}, 2016.

\bibitem{carvalho2015illuminant}
T.~Carvalho, F.~A. Faria, H.~Pedrini, R.~d.~S. Torres, and A.~Rocha.
\newblock Illuminant-based transformed spaces for image forensics.
\newblock {\em IEEE transactions on information forensics and security},
  11(4):720--733, 2015.

\bibitem{chesney2018deep}
R.~Chesney and D.~K. Citron.
\newblock Deep fakes: a looming challenge for privacy, democracy, and national
  security.
\newblock 2018.

\bibitem{de2013exposing}
T.~J. De~Carvalho, C.~Riess, E.~Angelopoulou, H.~Pedrini, and
  A.~de~Rezende~Rocha.
\newblock Exposing digital image forgeries by illumination color
  classification.
\newblock {\em IEEE Transactions on Information Forensics and Security},
  8(7):1182--1194, 2013.

\bibitem{dixit2017aga}
M.~Dixit, R.~Kwitt, M.~Niethammer, and N.~Vasconcelos.
\newblock Aga: Attribute-guided augmentation.
\newblock In {\em Proceedings of the IEEE Conference on Computer Vision and
  Pattern Recognition}, pages 7455--7463, 2017.

\bibitem{dolhansky2019deepfake}
B.~Dolhansky, R.~Howes, B.~Pflaum, N.~Baram, and C.~C. Ferrer.
\newblock The deepfake detection challenge (dfdc) preview dataset.
\newblock {\em arXiv preprint arXiv:1910.08854}, 2019.

\bibitem{fei2006one}
L.~Fei-Fei, R.~Fergus, and P.~Perona.
\newblock One-shot learning of object categories.
\newblock {\em IEEE transactions on pattern analysis and machine intelligence},
  28(4):594--611, 2006.

\bibitem{finn2017model}
C.~Finn, P.~Abbeel, and S.~Levine.
\newblock Model-agnostic meta-learning for fast adaptation of deep networks.
\newblock In {\em Proceedings of the 34th International Conference on Machine
  Learning-Volume 70}, pages 1126--1135. JMLR. org, 2017.

\bibitem{goodfellow2014generative}
I.~Goodfellow, J.~Pouget-Abadie, M.~Mirza, B.~Xu, D.~Warde-Farley, S.~Ozair,
  A.~Courville, and Y.~Bengio.
\newblock Generative adversarial nets.
\newblock In {\em Advances in neural information processing systems}, pages
  2672--2680, 2014.

\bibitem{gulrajani2017improved}
I.~Gulrajani, F.~Ahmed, M.~Arjovsky, V.~Dumoulin, and A.~C. Courville.
\newblock Improved training of wasserstein gans.
\newblock In {\em Advances in neural information processing systems}, pages
  5767--5777, 2017.

\bibitem{hariharan2017low}
B.~Hariharan and R.~Girshick.
\newblock Low-shot visual recognition by shrinking and hallucinating features.
\newblock In {\em Proceedings of the IEEE International Conference on Computer
  Vision}, pages 3018--3027, 2017.

\bibitem{he2017mask}
K.~He, G.~Gkioxari, P.~Doll{\'a}r, and R.~Girshick.
\newblock Mask r-cnn.
\newblock In {\em Proceedings of the IEEE international conference on computer
  vision}, pages 2961--2969, 2017.

\bibitem{he2016deep}
K.~He, X.~Zhang, S.~Ren, and J.~Sun.
\newblock Deep residual learning for image recognition.
\newblock In {\em Proceedings of the IEEE conference on computer vision and
  pattern recognition}, pages 770--778, 2016.

\bibitem{huang2017arbitrary}
X.~Huang and S.~Belongie.
\newblock Arbitrary style transfer in real-time with adaptive instance
  normalization.
\newblock In {\em Proceedings of the IEEE International Conference on Computer
  Vision}, pages 1501--1510, 2017.

\bibitem{karras2017progressive}
T.~Karras, T.~Aila, S.~Laine, and J.~Lehtinen.
\newblock Progressive growing of gans for improved quality, stability, and
  variation.
\newblock {\em arXiv preprint arXiv:1710.10196}, 2017.

\bibitem{karras2019style}
T.~Karras, S.~Laine, and T.~Aila.
\newblock A style-based generator architecture for generative adversarial
  networks.
\newblock In {\em Proceedings of the IEEE Conference on Computer Vision and
  Pattern Recognition}, pages 4401--4410, 2019.

\bibitem{kingma2013auto}
D.~P. Kingma and M.~Welling.
\newblock Auto-encoding variational bayes.
\newblock {\em arXiv preprint arXiv:1312.6114}, 2013.

\bibitem{korshunova2017fast}
I.~Korshunova, W.~Shi, J.~Dambre, and L.~Theis.
\newblock Fast face-swap using convolutional neural networks.
\newblock In {\em Proceedings of the IEEE International Conference on Computer
  Vision}, pages 3677--3685, 2017.

\bibitem{li2018ictu}
Y.~Li, M.-C. Chang, and S.~Lyu.
\newblock In ictu oculi: Exposing ai created fake videos by detecting eye
  blinking.
\newblock In {\em 2018 IEEE International Workshop on Information Forensics and
  Security (WIFS)}, pages 1--7. IEEE, 2018.

\bibitem{liu2018unified}
A.~H. Liu, Y.-C. Liu, Y.-Y. Yeh, and Y.-C.~F. Wang.
\newblock A unified feature disentangler for multi-domain image translation and
  manipulation.
\newblock In {\em Advances in Neural Information Processing Systems}, pages
  2590--2599, 2018.

\bibitem{liu2019few}
M.-Y. Liu, X.~Huang, A.~Mallya, T.~Karras, T.~Aila, J.~Lehtinen, and J.~Kautz.
\newblock Few-shot unsupervised image-to-image translation.
\newblock {\em arXiv preprint arXiv:1905.01723}, 2019.

\bibitem{liu2016ssd}
W.~Liu, D.~Anguelov, D.~Erhan, C.~Szegedy, S.~Reed, C.-Y. Fu, and A.~C. Berg.
\newblock Ssd: Single shot multibox detector.
\newblock In {\em European conference on computer vision}, pages 21--37.
  Springer, 2016.

\bibitem{lombardi2018deep}
S.~Lombardi, J.~Saragih, T.~Simon, and Y.~Sheikh.
\newblock Deep appearance models for face rendering.
\newblock {\em ACM Transactions on Graphics (TOG)}, 37(4):68, 2018.

\bibitem{motiian2017few}
S.~Motiian, Q.~Jones, S.~Iranmanesh, and G.~Doretto.
\newblock Few-shot adversarial domain adaptation.
\newblock In {\em Advances in Neural Information Processing Systems}, pages
  6670--6680, 2017.

\bibitem{motiian2017unified}
S.~Motiian, M.~Piccirilli, D.~A. Adjeroh, and G.~Doretto.
\newblock Unified deep supervised domain adaptation and generalization.
\newblock In {\em Proceedings of the IEEE International Conference on Computer
  Vision}, pages 5715--5725, 2017.

\bibitem{munkhdalai2017meta}
T.~Munkhdalai and H.~Yu.
\newblock Meta networks.
\newblock In {\em Proceedings of the 34th International Conference on Machine
  Learning-Volume 70}, pages 2554--2563. JMLR. org, 2017.

\bibitem{nagano2018pagan}
K.~Nagano, J.~Seo, J.~Xing, L.~Wei, Z.~Li, S.~Saito, A.~Agarwal, J.~Fursund,
  H.~Li, R.~Roberts, et~al.
\newblock pagan: real-time avatars using dynamic textures.
\newblock {\em ACM Trans. Graph.}, 37(6):258--1, 2018.

\bibitem{nichol2018first}
A.~Nichol, J.~Achiam, and J.~Schulman.
\newblock On first-order meta-learning algorithms.
\newblock {\em arXiv preprint arXiv:1803.02999}, 2018.

\bibitem{nirkin2019fsgan}
Y.~Nirkin, Y.~Keller, and T.~Hassner.
\newblock Fsgan: Subject agnostic face swapping and reenactment.
\newblock In {\em Proceedings of the IEEE International Conference on Computer
  Vision}, pages 7184--7193, 2019.

\bibitem{olszewski2017realistic}
K.~Olszewski, Z.~Li, C.~Yang, Y.~Zhou, R.~Yu, Z.~Huang, S.~Xiang, S.~Saito,
  P.~Kohli, and H.~Li.
\newblock Realistic dynamic facial textures from a single image using gans.
\newblock In {\em Proceedings of the IEEE International Conference on Computer
  Vision}, pages 5429--5438, 2017.

\bibitem{ravi2016optimization}
S.~Ravi and H.~Larochelle.
\newblock Optimization as a model for few-shot learning.
\newblock 2016.

\bibitem{ren2015faster}
S.~Ren, K.~He, R.~Girshick, and J.~Sun.
\newblock Faster r-cnn: Towards real-time object detection with region proposal
  networks.
\newblock In {\em Advances in neural information processing systems}, pages
  91--99, 2015.

\bibitem{rossler2019faceforensics++}
A.~R{\"o}ssler, D.~Cozzolino, L.~Verdoliva, C.~Riess, J.~Thies, and
  M.~Nie{\ss}ner.
\newblock Faceforensics++: Learning to detect manipulated facial images.
\newblock {\em arXiv preprint arXiv:1901.08971}, 2019.

\bibitem{salakhutdinov2012one}
R.~Salakhutdinov, J.~Tenenbaum, and A.~Torralba.
\newblock One-shot learning with a hierarchical nonparametric bayesian model.
\newblock In {\em Proceedings of ICML Workshop on Unsupervised and Transfer
  Learning}, pages 195--206, 2012.

\bibitem{snell2017prototypical}
J.~Snell, K.~Swersky, and R.~Zemel.
\newblock Prototypical networks for few-shot learning.
\newblock In {\em Advances in Neural Information Processing Systems}, pages
  4077--4087, 2017.

\bibitem{sung2018learning}
F.~Sung, Y.~Yang, L.~Zhang, T.~Xiang, P.~H. Torr, and T.~M. Hospedales.
\newblock Learning to compare: Relation network for few-shot learning.
\newblock In {\em Proceedings of the IEEE Conference on Computer Vision and
  Pattern Recognition}, pages 1199--1208, 2018.

\bibitem{suwajanakorn2017synthesizing}
S.~Suwajanakorn, S.~M. Seitz, and I.~Kemelmacher-Shlizerman.
\newblock Synthesizing obama: learning lip sync from audio.
\newblock {\em ACM Transactions on Graphics (TOG)}, 36(4):95, 2017.

\bibitem{thies2016face2face}
J.~Thies, M.~Zollhofer, M.~Stamminger, C.~Theobalt, and M.~Nie{\ss}ner.
\newblock Face2face: Real-time face capture and reenactment of rgb videos.
\newblock In {\em Proceedings of the IEEE Conference on Computer Vision and
  Pattern Recognition}, pages 2387--2395, 2016.

\bibitem{valverde2019one}
S.~Valverde, M.~Salem, M.~Cabezas, D.~Pareto, J.~C. Vilanova,
  L.~Rami{\'o}-Torrent{\`a}, {\`A}.~Rovira, J.~Salvi, A.~Oliver, and
  X.~Llad{\'o}.
\newblock One-shot domain adaptation in multiple sclerosis lesion segmentation
  using convolutional neural networks.
\newblock {\em NeuroImage: Clinical}, 21:101638, 2019.

\bibitem{vinyals2016matching}
O.~Vinyals, C.~Blundell, T.~Lillicrap, D.~Wierstra, et~al.
\newblock Matching networks for one shot learning.
\newblock In {\em Advances in neural information processing systems}, pages
  3630--3638, 2016.

\bibitem{wang2018low}
Y.-X. Wang, R.~Girshick, M.~Hebert, and B.~Hariharan.
\newblock Low-shot learning from imaginary data.
\newblock In {\em Proceedings of the IEEE Conference on Computer Vision and
  Pattern Recognition}, pages 7278--7286, 2018.

\bibitem{yang2019exposing}
X.~Yang, Y.~Li, and S.~Lyu.
\newblock Exposing deep fakes using inconsistent head poses.
\newblock In {\em ICASSP 2019-2019 IEEE International Conference on Acoustics,
  Speech and Signal Processing (ICASSP)}, pages 8261--8265. IEEE, 2019.

\bibitem{zakharov2019few}
E.~Zakharov, A.~Shysheya, E.~Burkov, and V.~Lempitsky.
\newblock Few-shot adversarial learning of realistic neural talking head
  models.
\newblock In {\em Proceedings of the IEEE International Conference on Computer
  Vision}, pages 9459--9468, 2019.

\bibitem{zhou2017two}
P.~Zhou, X.~Han, V.~I. Morariu, and L.~S. Davis.
\newblock Two-stream neural networks for tampered face detection.
\newblock In {\em 2017 IEEE Conference on Computer Vision and Pattern
  Recognition Workshops (CVPRW)}, pages 1831--1839. IEEE, 2017.

\bibitem{zhu2016generative}
J.-Y. Zhu, P.~Kr{\"a}henb{\"u}hl, E.~Shechtman, and A.~A. Efros.
\newblock Generative visual manipulation on the natural image manifold.
\newblock In {\em European Conference on Computer Vision}, pages 597--613.
  Springer, 2016.

\bibitem{zollhofer2018state}
M.~Zollh{\"o}fer, J.~Thies, P.~Garrido, D.~Bradley, T.~Beeler, P.~P{\'e}rez,
  M.~Stamminger, M.~Nie{\ss}ner, and C.~Theobalt.
\newblock State of the art on monocular 3d face reconstruction, tracking, and
  applications.
\newblock In {\em Computer Graphics Forum}, volume~37, pages 523--550. Wiley
  Online Library, 2018.

\end{thebibliography}
}

\end{document}